\documentclass[conference]{IEEEtran}
\IEEEoverridecommandlockouts

\usepackage{cite}
\usepackage{amsmath,amssymb,amsfonts}
\usepackage[ruled,vlined,linesnumbered]{algorithm2e}
\usepackage{graphicx}
\usepackage{textcomp}
\usepackage{xcolor}
\usepackage{balance}
\usepackage{multirow}
\usepackage{adjustbox}
\usepackage{booktabs}
\usepackage{multicol}
\usepackage{tabu}

\usepackage{threeparttable} 
\SetKwInput{KwInput}{Input} 
\SetKwInput{KwNull}{null}

\usepackage{xcolor}

\begin{document}
  

\title{GNN-ViTCap: GNN-Enhanced Multiple Instance Learning with Vision Transformers for Whole Slide Image Classification and Captioning}



\author{S M Taslim Uddin Raju\textsuperscript{1}, Md. Milon Islam\textsuperscript{1}, Md Rezwanul Haque\textsuperscript{1}, Hamdi Altaheri\textsuperscript{1}, and Fakhri Karray\textsuperscript{1,2}

\thanks{\textsuperscript{1}The authors are with the Centre for Pattern Analysis and Machine Intelligence, Department of Electrical and Computer Engineering, University of Waterloo, N2L 3G1, Ontario, Canada. (e-mail: smturaju@uwaterloo.ca{*}, milonislam@uwaterloo.ca, rezwan@uwaterloo.ca, haltaheri@uwaterloo.ca).

\textsuperscript{1,2}The author is with the Machine Learning Department at Mohamed bin Zayed University of Artificial Intelligence, Abu Dhabi, United Arab Emirates (email: fakhri.karray@mbzuai.ac.ae) and the Centre for Pattern Analysis and Machine Intelligence, Department of Electrical and Computer Engineering, University of Waterloo, N2L 3G1, Ontario, Canada  (e-mail: karray@uwaterloo.ca).}
}

\maketitle

\begin{abstract}

Microscopic assessment of histopathology images is vital for accurate cancer diagnosis and treatment. Whole Slide Image (WSI) classification and captioning have become crucial tasks in computer-aided pathology. However, microscopic WSIs face challenges such as redundant patches and unknown patch positions due to subjective pathologist captures. Moreover, generating automatic pathology captions remains a significant challenge. To address these challenges, a novel GNN-ViTCap framework is introduced for classification and caption generation from histopathological microscopic images. A visual feature extractor is used to extract feature embeddings. The redundant patches are then removed by dynamically clustering images using deep embedded clustering and extracting representative images through a scalar dot attention mechanism. The graph is formed by constructing edges from the similarity matrix, connecting each node to its nearest neighbors. Therefore, a graph neural network is utilized to extract and represent contextual information from both local and global areas. The aggregated image embeddings are then projected into the language model’s input space using a linear layer and combined with input caption tokens to fine-tune the large language models for caption generation. Our proposed method is validated using the BreakHis and PatchGastric microscopic datasets. The GNN-ViTCap method achieves an $F_1$-Score of 0.934 and AUC of 0.963 for classification, along with BLEU@4 = 0.811 and METEOR = 0.569 for captioning. Experimental analysis demonstrates that the GNN-ViTCap architecture outperforms state-of-the-art (SOTA) approaches, providing a reliable and efficient approach for patient diagnosis using microscopy images.
\end{abstract}
\section{Introduction}
Histopathology is the microscopic examination of tissue, which is the benchmark for cancer diagnosis and treatment decisions \cite{tsuneki2022inference}. Recently, the advancement of deep learning technologies has significantly propelled computational histopathology, particularly by training models with gigapixel Whole Slide Images (WSIs) obtained from Hematoxylin and Eosin (H\&E)-stained specimens \cite{song2023artificial}. Although cancer detection and classification are crucial, pathologists typically prepare diagnostic reports based on their observations of H\&E-stained slides. These detailed captions provide invaluable insights that enhance the diagnostic process. Therefore, automated pathological report generation can make model predictions more understandable and give the pathologists richer contextual information to guide their decision-making. Thus, collecting high-quality WSI-text pairs is crucial for advancing visual-language models and promoting innovation in computational pathology \cite{ahmed2024pathalign}.

However, histopathology WSIs have limitations due to their large size, complicating computational analysis. Multiple Instance Learning (MIL) has emerged as a popular approach for WSI classification by segmenting large images into smaller patches (i.e., instances) to form a slide-level (or bag-level) representation. However, MIL assumes patches from the same WSI are independent, neglecting vital tissue context and spatial interactions \cite{ilse2018attention}. In contrast, pathologists consider spatial organization and relationships between patches for comprehensive tissue analysis. Recently, microscopic imaging offers cost and memory advantages as an alternative to scanned WSIs. However, it introduces complexities such as the absence of absolute positional data and redundant patches from multiple subjective captures. Fig. 1(a) demonstrates the scanner WSI where the absolute position of the patches is known.
In contrast, Fig. 1(b) presents the microscopic WSI, which lacks the absolute position of patches and exhibits redundant patches due to multiple captures from a pathologist’s subjective perspective. Existing approaches such as ABMIL \cite{ilse2018attention}, and TransMIL \cite{shao2021transmil} struggle with global modeling and capturing long-distance dependencies due to limited consideration of inter-patches relationships. Graph-based MIL approaches \cite{ahmedt2022survey} also encounter challenges in microscopic image analysis within patients, as the absence of absolute patches prevents the construction of effective graph edges. Furthermore, the redundancy in microscopic images leads to excessively dense and repetitive graph connections, which reduces the models' ability to capture global context and long-distance dependencies effectively.

Moreover, recurrent architectures such as RNNs and LSTMs have become popular in the field of pathological image captioning due to their capacity to model sequential data \cite{elbedwehy2024enhanced, tsuneki2022inference}. However, these methods struggle with vanishing gradients and cannot resolve long-range dependencies. Recently, large language models (LLMs) have drawn significant attention for their advanced capability to process and understand complex text \cite{touvron2023llama}. Biomedical language models demonstrate outstanding ability at caption generation by effectively translating complex medical images into clinically relevant descriptions \cite{lu2022clinicalt5, luo2022biogpt, zhang2024generalist}. In addition, vision transformers (ViTs) leverage transformer-based architectures to transform visual data into high-dimensional representations, enhancing image analysis \cite{russakovsky2015imagenet}. Integrating ViTs with biomedical language models facilitates multimodal integration of visual and textual data, thereby enhancing caption generation and classification tasks for more accurate medical image interpretation \cite{radford2021learning, jia2021scaling, zhou2024pathm3}. 


To address the problems as mentioned earlier, the GNN-ViTCap method is proposed for classification and caption generation from histopathological microscopic images. The method comprises a visual extractor and the high capabilities of biomedical language models. An attention-based deep embedded clustering method selects the most representative images to remove the redundant images, and graph-based aggregation (GNN-MIL) leverages the spatial relationships between image patches. The large language model combined with visual feature embeddings exhibits exceptional context-association capabilities, offering a promising alternative to RNNs or LSTM-based methods. The contributions can be summarized as follows: \begin{itemize}
    \item Applying a visual feature extractor to extract feature embeddings.
    \item Developing an attention-based deep embedded clustering method and selecting representative images using a scaled dot attention mechanism to remove redundancy in microscopic images.
    \item Constructing the graph neural network based on the similarity of representative images to capture spatial and contextual information within and between clusters.
    \item Projecting aggregated image embeddings into the language model's input space using a linear layer, then combining them with caption tokens to fine-tune the LLMs for caption generation.
\end{itemize}
The rest of this paper is summarized as follows: Section II reviews existing works, including current literature on WSI classification and captioning. Section III demonstrates the details of the proposed GNN-ViTCap architecture, including feature extraction, self-attention-based deep embedded clustering, GNN-MIL, and large language models.  Section IV describes the microscopic WSI datasets and experimental setup of the proposed method. Section V provides the research questions along with comprehensive evaluations and experimental results. Lastly, Section VI sums up the paper with a review of findings and directions for future research.

\section{Related Works}
\label{Literature Review_stage}
\subsection{Multiple Instance Learning in Histopathology}
MIL is now a significant method for analyzing WSIs with pyramid structure and giga-pixel dimensions in digital pathology, highlighting the applications of cancer subtyping, staging, grading, and tissue segmentation. The primary challenge associated with MIL is to combine the large number of instance features to generate a comprehensive bag feature for the specific task. Ilse et al. \cite{ilse2018attention} introduced an attention-based aggregation operator (ABMIL) utilizing a learnable neural network to enhance the contribution of each instance via trainable attention weights. A Dual-Stream MIL (DSMIL) is proposed in \cite{li2021dual} to classify the WSI using multi-scale features. In this network, all high-magnification patches are fused with their respective low-magnification samples, leading to considerable data redundancy. In another research, a Transformer-based MIL (TransMIL) \cite{shao2021transmil} is developed to investigate both morphological and spatial data in WSI classification. TransMIL leveraged the self-attention technique, utilizing the output data of a transformer network to encode the mutual correlations among instances. A Double-Tier Feature Distillation Multiple Instance Learning (DTFD-MIL) \cite{zhang2022dtfd} is designed to increase the total number of bags by utilizing pseudo-bags, resulting in a double-tier MIL system that effectively leverages inherent features. This network determined the probability of the instance within the attention-based MIL framework and employed this derivation to assist in generating and analyzing image features.
\subsection{Vision Language Pretraining} 
Recent studies have focused on pre-training multimodal models that combine visual and textual data to analyze histopathology images. A few works, including CLIP \cite{radford2021learning} and ALIGN \cite{jia2021scaling}, have shown that training on large and various web-sourced datasets of paired image samples and captions enables networks to build excellent zero-shot transfer abilities. These models attain this by employing prompts that deploy cross-modal alignment between visuals and texts learned during the pre-training phase. MI-Zero \cite{lu2023visual} is introduced to leverage the zero-shot transfer abilities of contrastively correlated visual and text architectures for histopathological WSI samples. This system enabled pretrained encoders to execute various downstream diagnostic tasks without requiring extra labeling. Moreover, the MI-Zero architecture redefines zero-shot transfer within the MIL framework, resolving the computational difficulties of performing inference on very large images. In PathAlign \cite{ahmed2024pathalign}, the use of BLIP-2 improved image-text alignment for giga-pixel WSIs. This framework obtained vision-language alignment by linking WSIs with their associated medical texts from pathology records. The pre-trained LLM and the WSI encoder are aligned in this research to create efficient and significant diagnostic reports from WSIs. CPath-Omni \cite{sun2024cpath} was introduced for patch and WSI level image analysis, such as classification and captioning that contained 15 billion LLM parameters. CPath-CLIP, a CLIP-based visual processor, is designed in this architecture to combine various vision models and include an LLM as a text encoder to develop a more robust CLIP framework.

\begin{figure*}[!ht]
    \centering
\centerline{\includegraphics[width=1\textwidth,height=0.62\textheight]{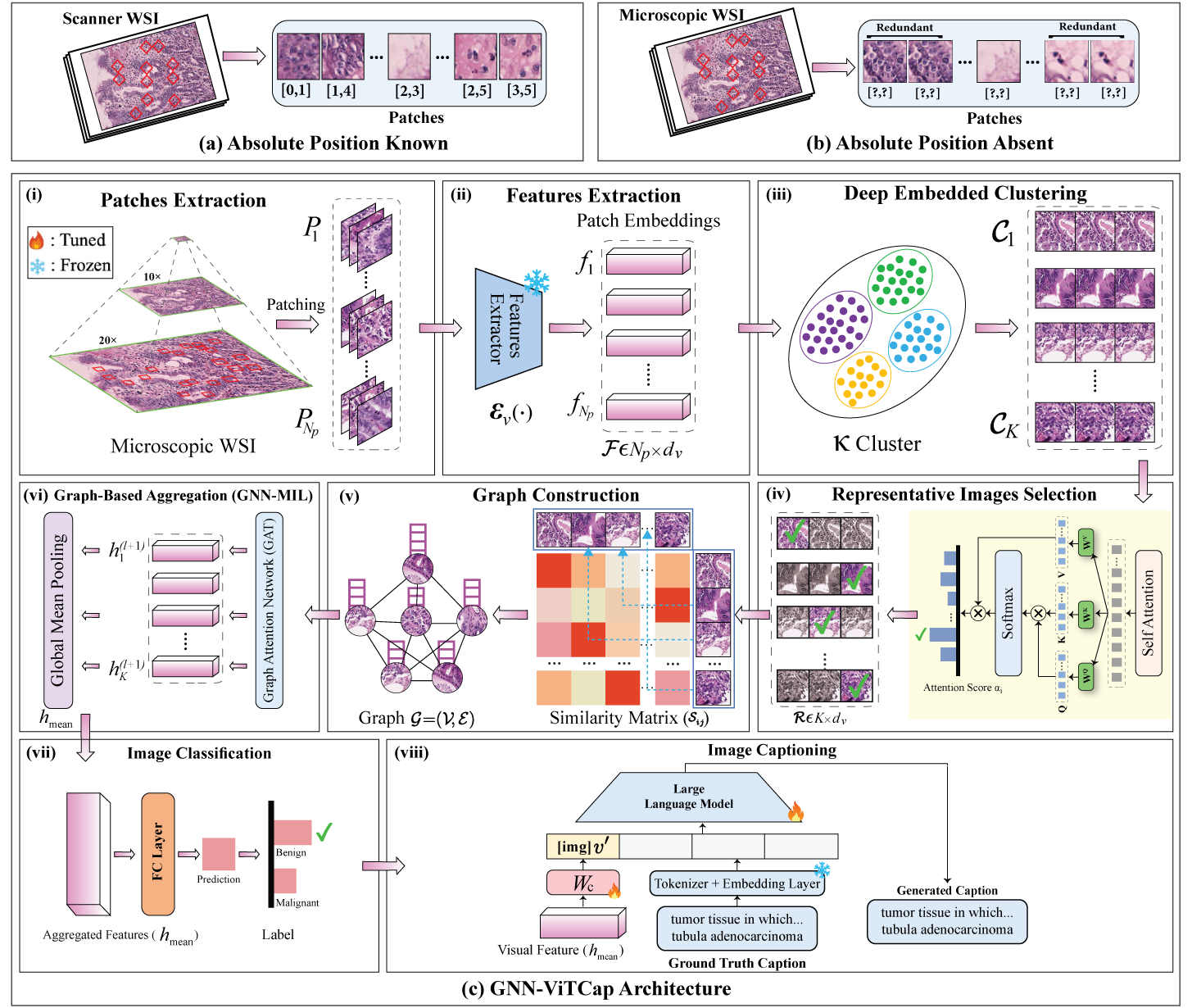}}
\vspace{-2mm}
\caption{\textbf{Overview of the GNN-ViTCap framework for microscopic whole slide images classification and captioning.} \textbf{(a)} Scanner WSI where the absolute position of each patch is known. \textbf{(b)} Microscopic WSI lacks patch position and contains redundant patches due to multiple captures from a pathologist’s subjective perspective. \textbf{(c) GNN-ViTCap architecture:} (i) extracting the patches from whole slide image, (ii) extracting image embeddings using a visual feature extractor, (iii) removing redundancy through deep embedded clustering, (iv) extracting representative images with scalar dot attention mechanism, (v) constructing a graph neural network (GNN) using the similarity of representative patches to capture contextual information within clusters (local) and between different clusters (regional), (vi) applying global mean pooling, which aggregates all node representations, (vii) classifying microscopic WSI using aggregated image embeddings and, (viii) projecting the aggregated image embeddings into the language model’s input space using a linear layer, and combining these projections with input caption tokens fine-tunes the LLMs for caption generation.}
\label{system}
\end{figure*}
\subsection{Biomedical Language Model for Image Captioning}
Biomedical language models have become crucial methods for generating informative captions from histopathological images, improving the interpretability and usability of visual data in healthcare environments \cite{nazi2024large}. These architectures utilize big pre-trained language models that have been fine-tuned using biomedical literature and annotated image datasets to generate relevant explanations of WSIs. The work presented in \cite{tsuneki2022inference} trained a baseline attention-based architecture that included a Convolutional Neural Network (CNN) encoder to extract significant features and an RNN decoder to predict captions based on features generated from patches. Qin et al. \cite{qin2023whole} developed a Subtype-guided Masked Transformer (SGMT) network to generate captions using transformers that used WSI samples as an input sequence and created caption sentences based on the input patches. A multimodal multi-task MIL system called PathM3 \cite{zhou2024pathm3} was proposed for WSI to classify and generate captions. PathM3 uses a query-based transformer to accurately correlate WSIs with diagnostic texts, even while training multi-task joint learning with minimal text data.

\section{Method: GNN-ViTCap Architecture}
Fig. 1(c) demonstrates the proposed GNN-ViTCap architecture for classification and captioning from microscopic whole slide images.

\subsection{MIL Formulation}
Multiple Instance Learning organizes data into \textit{bags} containing multiple \textit{instances}, with labels provided only at the bag level. MIL can be divided into two approaches: instance-based models and bag embedding-based models. Bag embedding-based models are preferred for their richer representations, enhancing WSI classification effectiveness \cite{shao2021transmil}. WSI classification is formulated as an MIL problem, where each slide is considered a \textit{bag} and its patches are \textit{instances}. Consider a binary classification problem with a dataset $\mathcal{D} = \{X_i, Y_i\}_{i=1}^{B}$ where $B$ is the total number of bags. Each bag $X_i$ is represented as a set of instances $X_i = \{x_{i,1}, x_{i,2}, \dots, x_{i,n_i}\}$, where $n_i$ is the number of instances in the $i^{th}$ bag. The label $Y_i \in \{0,1\}$ is assigned to the bag as a whole rather than to individual instances. According to the standard MIL assumption, the bag label $Y_i$ is described as: 

\begin{equation}
\centering
Y_i = \begin{cases} 
 0 & \text{iif} \sum_{j= 1}^{n_i} y_{i,j} =0 \\
1, & \text{otherwise}
\end{cases}
\label{bag}
\end{equation}
which can be modeled using max-pooling \cite{hou2016patch}. In another approach, the bag label $\hat{Y_i}$ can be determined using an aggregation function followed by a classifier as:
\begin{equation}
\hat{Y_i} = g\left(\sigma_{\text{AvgPool}} \left(f(x_{i,1}), f(x_{i,2}), \dots, f(x_{i,n_i})\right)\right)
\label{g(fx)}
\end{equation}
where $f(\cdot)$ is an instance-level feature extractor, $\sigma_\text{AvgPool}(\cdot)$ is a permutation-invariant function such as MIL pooling operation, and $g(\cdot)$ is the classifier to predict the bag label \cite{ilse2018attention}.

\subsection{Large Language Models}
The integral component of the proposed method involves utilizing both encoder-decoder and decoder-only language model architectures, specifically designed for autoregressive text generation. The image caption \( C = C_1, \dots, C_T \) is tokenized into multiple tokens and passed through an embedded layer to generate a sequence of embeddings. These embeddings are then fed to the decoder, which produces output embeddings. A linear classifier predicts the next token based on these outputs. The LLMs are trained by minimizing the cross-entropy loss between the predicted tokens and the ground truth. To address the caption generation task, GNN-ViTCap utilizes a combination of four LLMs: ClinicalT5-Base \cite{lu2022clinicalt5}, employed as an encoder-decoder model, and BioGPT \cite{luo2022biogpt}, LLamaV2-Chat \cite{touvron2023llama}, and BiomedGPT \cite{zhang2024generalist}, which are utilized as decoder-only models. These models effectively capture linguistic structures and generate coherent text.


\subsection{Vision Encodings}  
The GNN-ViTCap integrates a vision encoder with language models to simultaneously process visual and text inputs. For each patient $s$, given a microscopic whole slide image $X^{(s)}$, associated with the label $Y^{(s)}$ and the corresponding caption $C^{(s)}$. The WSI $X^{(s)}$ comprises a collection of $N_p$ patches or images, $\mathcal{P}^{(s)} = \{ p_k^{(s)} \}_{k=1}^{N_p}$, where $p_k^{(s)}$ denotes the $k$-th patch of patient $s$. The number of patches $N_p$ varies depending on the specific WSI and the individual characteristics of patient $s$. The pretrained vision encoder $\mathcal{E}_v$ processes each patch $p_k^{(s)}$ to produce a fixed-length embedding feature vector, a.k.a., a patch token:
\begin{equation}
f_{k}^{(s)} = \mathcal{E}_v(p_k^{(s)}) \in \mathbb{R}^{1 \times d_v}
    \label{z_v}
\end{equation}
where, $d_v$ is the dimension of the visual embeddings. The patch-level embeddings are concatenated to represent the entire WSI of each patient $s$:
\begin{equation}
\mathcal{F}_{(s)} = \left[f_1^{(s)}; f_2^{(s)}; \dots; f_{N_p}^{(s)}\right] \in \mathbb{R}^{N_p \times d_v}
\label{z_c}
\end{equation}

\subsection{Attention-based Deep Embedded Clustering}

In the proposed architecture, an attention-based deep embedded clustering (DEC) \cite{xie2016unsupervised} is utilized to remove the redundant images from each patient.
\subsubsection{Deep Embedded Clustering}
Let $F_{(s)} = \left\{f_1^{(s)}, f_2^{(s)}, \cdots, f_{N_p}^{(s)}\right\}$ denotes the set of features extracted from the images of patient $s$. The initial cluster centroids \( \mu^{(s)} = \{ \mu_k^{(s)} \}_{k=1}^K \) are used to assign these features into \( K \) image clusters, where \( K \) denotes the total number of clusters. The probability of assigning each feature embedding \( f_i^{(s)} \) to cluster \( k \) is determined using the Student’s \( t \)-distribution as follows: 
\begin{equation}
q_{ik}^{(s)} = \frac{\left( 1 + \| f_i^{(s)} - \mu_k^{(s)} \|^2/\alpha \right)^{- \frac{\alpha+1}{2}}}{\sum_{j=1}^K \left( 1 + \| f_i^{(s)} - \mu_j^{(s)} \|^2/\alpha \right)^{- \frac{\alpha+1}{2}}}
  \label{q_ik}
\end{equation}
where $\alpha = 1$ indicates the degree of freedom of Student’s \( t \)-distribution, and \( q_{ik}^{(s)} \) denotes the soft assignment probability of the \( i \)-th image from patient \( s \) to the \( k \)-th cluster. The higher value of \( q_{ik}^{(s)} \) indicates a stronger likelihood that the feature embedding \( f_i^{(s)} \) belongs to cluster \( k \). Therefore, an auxiliary target distribution \( \mathcal{T}_{(s)} = \{ t_{ik}^{(s)} \} \) is introduced based on the soft assignments \( \mathcal{Q}_{(s)} = \{ q_{ik}^{(s)} \} \) to refine the clustering process. The target distribution emphasizes high-confidence assignments and is computed as:
\begin{equation}
t_{ik}^{(s)} = \frac{ \left( q_{ik}^{(s)} \right)^2 / \sum_{i=1}^{N_p} q_{ik}^{(s)} }{ \sum_{j=1}^K \left( q_{ij}^{(s)} / \sum_{i=1}^{N_p} q_{ij}^{(s)} \right)^2 }
\label{t_ik}
\end{equation}
where, \( \sum_{i=1}^{N_p} q_{ik}(s) \) prevents bias toward larger clusters by balancing the influence of clusters with more assignments. The Kullback-Leibler (KL) divergence between the target distribution $\mathcal{T}_{(s)}$ and the soft assignment distribution $\mathcal{Q}_{(s)}$ is used as the clustering loss.
\begin{equation}
\mathcal{L}_{\text{Clu}} = \mathcal{KL} (\mathcal{T}_{(s)}|| \mathcal{Q}_{(s)}) =\sum_{i=1}^{N_p} \sum_{k=1}^K t_{ik}^{(s)} \log \left( \frac{t_{ik}^{(s)}}{q_{ik}^{(s)}} \right)
\label{KL}
\end{equation}
The objective of clustering is to minimize the distance between feature embeddings and their cluster centroids. By iteratively minimizing $\mathcal{L}_{\text{clu}}$, the clustering process refines both the embeddings and cluster centroids, resulting in compact and well-separated clusters.
\subsubsection{Attention-Based Representative Images}
DEC is applied to perform image clustering and capture relationships between images within the same area. To address redundancy and select the most representative image in each cluster, an attention-based selection mechanism is utilized \cite{vaswani2017attention}. This approach reduces intra-cluster redundancy while enhancing inter-cluster communication and information sharing. Let \(\mathcal{C}_k\) denotes the set of image indices corresponding to cluster \(k\). Therefore, each cluster \(k\) contains \(N_k = |\mathcal{C}_k|\) images, and the feature embeddings of these images are extracted as follows:
\begin{equation}
\mathcal{Z}_k = \{ f_i^{(s)} \mid i \in \mathcal{C}_k \} \in \mathbb{R}^{N_k \times d_v}
\label{Z_k}
\end{equation}
Then, the embeddings $\mathcal{Z}_k$ are mapped into query, key, and value representations using learnable linear projections.
\begin{equation}
Q^{\text{attn}} = \mathcal{Z}_k \mathcal{W}^Q_{\text{attn}}, \quad K^{\text{attn}} = \mathcal{Z}_k \mathcal{W}^K_{\text{attn}}, \quad V^{\text{attn}} = \mathcal{Z}_k \mathcal{W}^V_{\text{attn}}
\label{QKV_k_att}
\end{equation}
where \( \mathcal{W}^Q_{\text{attn}}, \mathcal{W}^K_{\text{attn}}, \mathcal{W}^V_{\text{attn}} \in \mathbb{R}^{d_v \times d_v} \) are parameter matrices. For each patch \( i \in \mathcal{C}_k \), an attention score is defined based on the element-wise product of its query \( Q_i^\text{attn} \in \mathbb{R}^{d_v} \) and key \( K_i^\text{attn} \in \mathbb{R}^{d_v} \):
\begin{equation}
e_i = \frac{\sum (Q_i^\text{attn} \odot K_i^\text{attn})}{\sqrt{d_v}}, \quad i=1,\dots,N_k.
\label{e_i}
\end{equation}
where, \( \sqrt{d_v} \) is a scaling factor to ensure balanced weight distributions. Therefore, the softmax function is applied across all patches in the cluster to obtain normalized attention weights:
\begin{equation}
\alpha_i = \frac{\exp(e_i)}{\sum_{j=1}^{N_k} \exp(e_j)}, \quad i = 1, \dots, N_k
\label{alpha_i}
\end{equation}
These weights \( \alpha_i \in \mathbb{R} \) indicate the relative significance of each patch \( i \) within the same cluster. Therefore, a scalar importance score is computed for each patch using its corresponding value embedding \( V_i^\text{attn} \in \mathbb{R}^{d_v} \):
\begin{equation}
\text{score}_i = \sum_{m=1}^{d_v} \alpha_i \cdot V_{i,m}^\text{attn}
\label{score_i}
\end{equation}
Furthermore, the image with the highest score is selected as the representative image within cluster $k$, and its corresponding embedding is defined as:
\begin{equation}
r_k^{(s)} = f_{i_k^*}^{(s)} \in \mathbb{R}^{d_v} \quad \text{where} \quad i_k^* = \arg\max_{i \in \mathcal{C}_k} \, \text{score}_i
\label{final_i}
\end{equation}
Therefore, the final representative features for patient $s$ is computed as follows:
\begin{equation}
\mathcal{R}_{(s)} = [r_1^{(s)}, r_1^{(s)}; \dots; r_K^{(s)}]^T \in \mathbb{R}^{K \times d_v}
\label{R_s}
\end{equation}
where $K$ is the total number of clusters. 
\subsection{Graph-Based Aggregation (GNN-MIL)}
Scanner-based WSI provides defined absolute positions and coordinates, which facilitate the identification of neighboring patches. In contrast, microscopy-based WSI lacks absolute coordinates, making it more challenging to determine neighboring patches. To address this, a graph-based aggregation approach is utilized to identify neighboring representative images based on their relative positions, thereby maximizing their interactions.
\subsubsection{Constructing Graph}
For each patient \( s \), the graph \( \mathcal{G}^{(s)} \) is constructed using a similarity matrix \( \mathcal{S}^{(s)} \in \mathbb{R}^{K \times K} \), where each node denotes a representative image. The similarity between nodes is calculated using cosine similarity as:
\begin{equation}
\mathcal{S}^{(s)}_{ij} = \langle \frac{r_i^{(s)}}{||r_i^{(s)}||_2}, \frac{r_j^{(s)}}{||r_j^{(s)}||_2} \rangle, \quad \forall i, j \in \{1, \dots, K\},
\label{S_s}
\end{equation}
where each element \( \mathcal{S}^{(s)}_{i,j} \) represents the similarity between image features \( r^{(s)}_i \) and \( r^{(s)}_j \). Therefore, an edge matrix $\mathcal{E}_{i,j}^{(p)}$ is created by applying the Gumbel Softmax function $\sigma_{\text{gsf}}$ \cite{jang2016categorical} to the entire similarity matrix $\mathcal{S}^{(s)}_{ij}$. This process selects the most similar neighbors for each node, resulting in: 
\begin{equation}
\mathcal{E}^{(s)}_{i,j} =
\begin{cases}
1, & \text{if } \mathcal{S}^{(s)}_{i,j} = \max \limits_{k \in \mathcal{N}(i)} \sigma_{\text{gsf}}(\mathcal{S}^{(s)}_{k,j}) \\
0, & \text{otherwise}
\end{cases}
\label{E_s}
\end{equation}
where \( \mathcal{E}_{i,j}^{(s)} \in \{0,1\} \) indicates the presence of an edge between node \( i \) and node \( j \) and $\mathcal{N}(i)$ denotes the set of neighbors for node $i$. Therefore, the graph \( \mathcal{G}_{(s)} \) is defined by its set of nodes \( \mathcal{V}_{(s)} \) and edges \( \mathcal{E}_{(s)} \):
\begin{equation}
\mathcal{G}_{(s)} = (\mathcal{V}_{(s)}, \mathcal{E}_{(s)})
\label{G_p}
\end{equation}
where \( \mathcal{V}_{(s)} = \mathcal{R}_{(s)} \) denotes the set of nodes corresponding to the representative images for patient \( s \), and \( \mathcal{E}_{(s)} \subseteq \mathcal{V}_{(s)} \times \mathcal{V}_{(s)} \) are edges determined by \( \mathcal{E}^{(s)}_{i,j} \).
\subsubsection{Graph Neural Network (GNN)}
In the graph $\mathcal{G}^{(s)}$, each node corresponds to a representative image, and the edges denote the relationships between these images. Graph Attention Networks (GAT) \cite{velickovic2017graph} improve attention mechanisms to assign adaptive weights to neighboring nodes during aggregation. For each layer \( l = 0, 1, \dots, L - 1 \), the features of node $v$ can be updated as follows: 
\begin{equation}
h_v^{(l+1)(s)} = \rho \left(\sum_{u \in \mathcal{N}(v)} \beta_{vu}^{(l)(s)} \mathcal{W}^{(l)} h_u^{(l)(s)}\right)
\label{h_v}
\end{equation}
where, $\mathcal{W}^{(l)} \in ^{d_v \times d_v}$ is the weight matrix at layer $l$ and $\rho$ is an activation function, such as LeakyReLU \cite{maas2013rectifier}. The attention coefficient \(\beta_{vu}^{(l)}\) between nodes \(u\) and \(v\) at layer \(l\) is computed as:
\begin{footnotesize}
\begin{equation}
\beta_{vu}^{(l)(s)} = \frac{\exp \left( \rho \left( a^{(l)^T} \left[ \mathcal{W}^{(l)} h_v^{(l)(s)} \| \mathcal{W}^{(l)} h_u^{(l)(s)} \right] \right) \right)}{\sum\limits_{w \in \mathcal{N}(v)} \exp \left( \rho \left( a^{(l)^T} \left[ \mathcal{W}^{(l)} h_v^{(l)(s)} \| \mathcal{W}^{(l)} h_w^{(l)(s)} \right] \right) \right)} 
\label{a_uv}
\end{equation}
\end{footnotesize}where \( a^{(l)} \) denotes a learnable weight vector at layer \( l \). After \( L \) GAT layers, the final node representations \( h_v^{(L)} \) are obtained, where each \( h_v^{(L)} \in \mathbb{R}^{d_{\text{out}}} \). Consequently, the WSI representation \( {h}_{\text{mean}}^{(s)} \in \mathbb{R}^{d_{\text{out}}}\) is generated by applying global mean pooling, which aggregates all node representations:
\begin{equation}
{h}_{\text{mean}}^{(s)}= \frac{1}{K} \sum_{v=1}^K h_v^{(L)(s)}
\label{h_bar}
\end{equation}

\subsection{Visual Embedding Projection}
The visual embeddings generated by the vision encoder are not directly compatible with the language model. This incompatibility arises because the vision embeddings have different dimensions and feature distributions compared to the language model's input space. To address this, the aggregated image embeddings \( h_{\text{mean}}^{(s)} \in \mathbb{R}^{d_{\text{out}}} \), are transformed using a linear projection matrix \( W_c \in \mathbb{R}^{d_{\text{out}} \times d_{\text{model}}} \). This projection maps the visual embeddings into a \( d_{\text{model}} \)-dimensional space compatible with the language model’s input embeddings. The visual prefix can be computed as: 
\begin{equation}
 v'_{(s)} = h_{\text{mean}}^{(s)} \cdot \mathcal{W}_c \in \mathbb{R}^{d_{\text{model}}}
\label{v'}
\end{equation}
Therefore, the visual prefix is integrated with the input caption token embeddings to fine-tune the language model for caption generation.

\subsection{Traninging Procedure}
The GNN-ViTCap architecture is trained for two tasks: image classification and image captioning.
\subsubsection{Image Classification} 
For the image classification, the aggregated image embeddings $h^{(s)}_{\text{mean}}$, are fed into multilayer perceptron (MLP) to predict the target variable:
\begin{equation}
\hat{y}_{(s)} = \text{MLP}(h_{\text{mean}}^{(s)})
\label{mlp_prediction}
\end{equation}
where, $\hat{y}_{(s)}$ represents the predicted probability for patient $s$. The binary cross-entropy loss function is then used for optimization and is defined as:
\begin{equation}
\mathcal{L}_{\text{BCE}} = -\frac{1}{N} \sum_{i=1}^{N} \left[ y_i \cdot \log\hat{y}_i + (1 - y_i) \cdot \log(1 - \hat{y}_i) \right]
\label{L_bcl}
\end{equation}
where, $N$ is the number of patients, $y_i$ is the ground-truth label, and $\hat{y}_i$ is the predicted probability for each patient. Therefore, the total loss for the whole slide image classification can be characterized as: 
\begin{equation}
    \mathcal{L}_{\text{Total}} = \mathcal{L}_{\text{BCL}} + \mathcal{L}_{\text{Clu}}
    \label{total_clss}
\end{equation}

\subsubsection{Caption Generation}
In the caption generation task, the visual prefix \( v'\), combined with the start-of-sequence token embeddings of the caption, is fed into the language model. The language model then autoregressively generates caption tokens \( C_t \) based on \( v' \) and the previously generated tokens \( C_1 \) to \( C_{t-1} \). The loss for caption generation is calculated using the negative log-likelihood of the ground-truth captions:
\begin{equation}
\mathcal{L}_{\text{Cap}} = -\frac{1}{N} \sum_{i=1}^{N} \sum_{t=1}^{T} \log p_{\theta}(C_{i,t} \mid v_i', C_{i,1}, \dots, C_{i,t-1})
\label{L_cap}
\end{equation}
where, \( T \) is the maximum caption length, \( C_{i,t} \) is the \( t \)-th token of the ground-truth caption for patient \(i \), and \( p_{\theta} \) represents the probability predicted by the language model with parameters \( \theta \). Therefore, the total loss for caption generation for the whole slide image can be characterized as: 
\begin{equation}
    \mathcal{L}_{\text{Total}} = \mathcal{L}_{\text{Cap}} + \mathcal{L}_{\text{Clu}}
    \label{total_cap}
\end{equation}

\section{Experimental Setup}
\subsection{Datasets}
\subsubsection{BreakHis}
The BreakHis dataset \cite{spanhol2015dataset} comprises 7,909 microscopic histopathology biopsy images from 82 patients. Each image is classified into benign and malignant tumor categories. In this study, labels are assigned only at the patient level within the dataset, without annotations for each image patch. The sample images are obtained from breast tissue biopsy slides stained with H\&E. The images are captured in RGB TrueColor with a 24-bit color depth (8 bits per channel) using magnifying factors of $40\times$, $100\times$, $200\times$, and $400\times$. Each image is stored in an uncompressed graphics format with dimensions of $700 \times 460$ pixels.
\subsubsection{PatchGastric}
The PatchGastric dataset \cite{tsuneki2022inference} consists of paired entries, each comprised of image patches from a stomach adenocarcinoma endoscopic biopsy specimen and corresponding histopathological caption. The image patches are obtained from whole slide images and associated with captions from diagnostic reports. There are 262,777 patches of dimension $300\times 300 $ pixels obtained from 991 H\&E-stained slides. Each slide is unique to an individual patient and is captured at $20\times$ magnification. The captions associated with the patients are composed of a vocabulary of 344 unique words, with each sentence containing up to 47 words.



\subsection{Implementation Details and Training Phase}
For feature extraction, an ImageNet-21k pre-trained ViT-B/16 \cite{russakovsky2015imagenet} is used as the visual backbone, which has $12-$layers of Vision Transformer that encodes images into embeddings of dimension $d_v = 768$. Images are processed at a resolution of $224\times224$ pixels with $16\times16$ patch size. Additionally, a ResNet-34 pre-trained on ImageNet with 34 convolutional layers produces embeddings of dimension $d_v = 512$. For attention-based deep embedded clustering, the number of clusters is set to $\mathcal{K} = 8$ for the BreakHis dataset and $\mathcal{K} = 50$ for the PatchGastric dataset, with a convergence threshold of $\epsilon = 10^{-4}$. The hidden layer dimension of the graph neural network is $d_\text{out} = 512$ with GAT layer $L = 3$. A linear projection layer with a dimension of $d_\text{model} = 768$ maps image features into the large language model input space. The tokenizer functions of our chosen LLMs facilitated the conversion of text into a format suitable for model processing. All methods are trained using cross-entropy loss with a learning rate of $10^{-3}$ and a dropout rate of $0.3$. The training is set to run for 100 epochs using the Adam optimizer, a weight decay of $10^{-2}$, and a batch size of $16$ for both the training and evaluation phases. The proposed architecture is implemented using PyTorch and the Deep Graph Library (DGL). Experiments are performed on an NVIDIA RTX $A6000$ graphics card GPU with $48$ GB of memory.
\subsection{Evaluation Metrics }
The evaluation metrics for WSI image classification include AUC (Area Under the Curve), Precision (Pr), Recall (Rc), and $F_1-$Score ($F_1)$. These metrics provide a comprehensive evaluation of the overall performance of the GNN-ViTCap architecture. The $F_1-$Score is particularly important for imbalanced datasets as it balances precision and recall, ensuring robust evaluation. AUC measures the model’s ability to distinguish between positive and negative cases, with higher scores indicating superior discrimination. In contrast, the GNN-ViTCap architecture is also used to generate text reports using information extracted from histopathological patches. For image captioning tasks, evaluation metrics such as BLEU, METEOR, ROUGE, and CIDEr are utilized \cite{sai2022survey}.
\section{Results and Discussions}
The main objective of this experiment is to explore the following questions: 
\begin{itemize}
    \item $Q_1$: \textit{Does the proposed GNN-MIL perform better than SOTA MIL methods for microscopic WSI classification?}
    \item $Q_2$: \textit{Does the spatial positional information of patches impact the performance of model for caption generation?}
    \item $Q_3$: \textit{Do LLMs perform better than LSTM or traditional transformer models for image captioning of WSI?}
    \item $Q_4$: \textit{Do in-domain LLMs perform better than generalized LLMs for generating captions in histopathological image analysis?}
\end{itemize}

\begin{table}[!bp]
\centering
  \vspace{-1mm}
\caption{Performance of GNN-ViTCap against SOTA methods on the BreakHis test dataset for classification.}
\vspace{-2mm}
\label{cls_com}
\renewcommand{\arraystretch}{1.25}
 \begin{threeparttable}
\begin{tabular}{l l l l l }
\hline 
Methods & Pr& Rc & $F_1$ & AUC \\ 
\hline
ABMIL\cite{ilse2018attention} & $0.835$ & $0.922$ & $0.900$ & $0.871$ \\
\hline
DSMIL \cite{li2021dual} & $0.872$ & $ 0.842$ & $0.856$ & $0.869 $\\
\hline
TransMIL\cite{shao2021transmil} &$0.865$ & $0.908$& $0.886$ &$0.862$ \\
\hline
DTFD-MIL \cite{zhang2022dtfd} & $0.854$ & $0.925$  & $0.911$ & $0.887$\\
\hline
GNN-ViTCap (ResNet-34) & $ 0.917$ &  $0.925$ & $0.921$& $0.906$\\
GNN-ViTCap (ViT-B/16)& \textbf{0.926} & \textbf{0.942} & \textbf{0.934} & \textbf{0.963}\\
\hline
\end{tabular}
    \end{threeparttable}
\end{table}
\subsection{Image Classification Results} 
\begin{figure}[!bp]
  \centering
       \vspace{-2mm}
  \begin{tabular}{ c  c }
    \begin{minipage}{.25\textwidth}
      \hspace{-.3cm}
      \includegraphics[width=\linewidth, height=38mm]{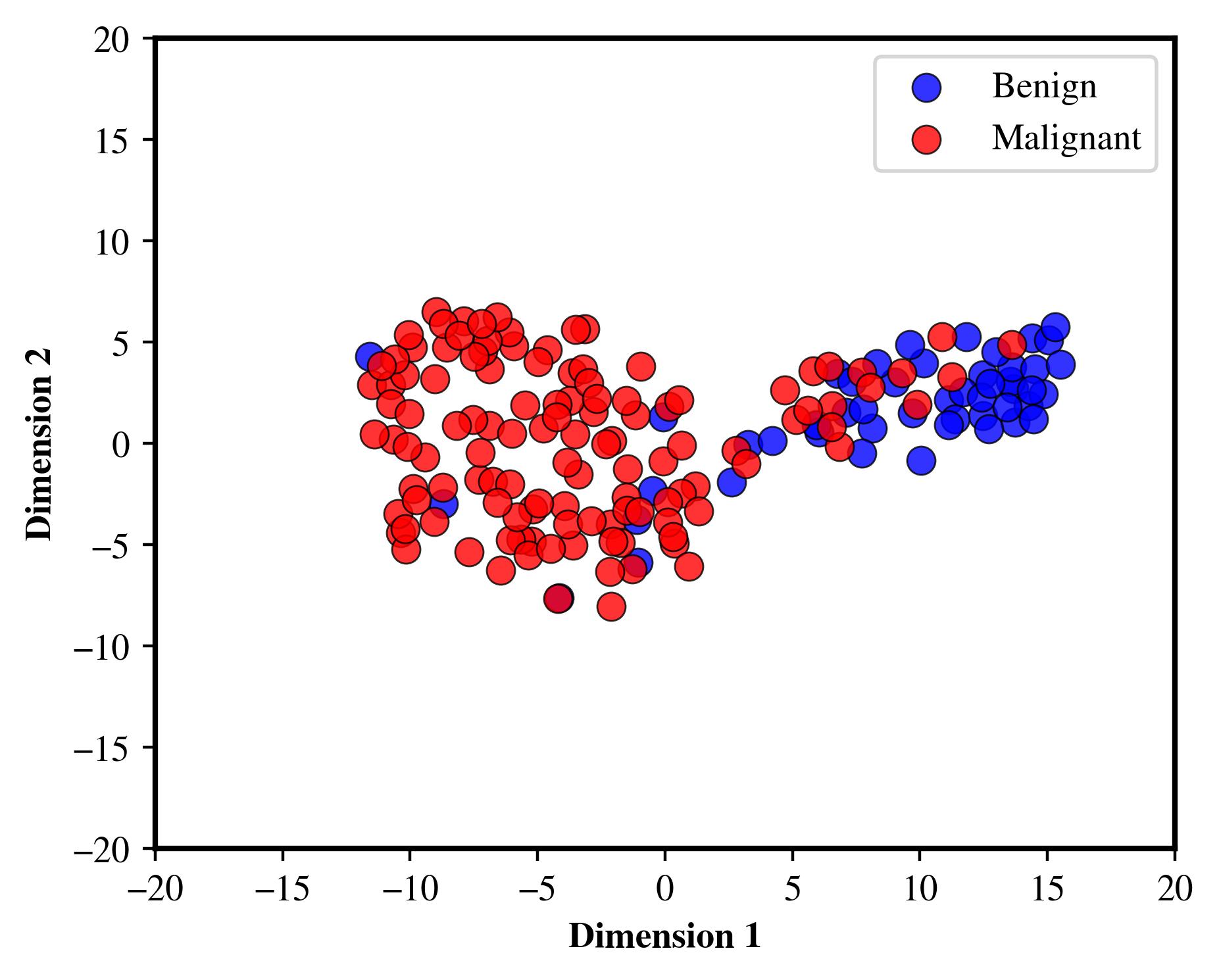}
    \end{minipage}
  \hspace{-.8cm}
    &
  \begin{minipage}{.25\textwidth}
      \includegraphics[width=\linewidth, height=38mm]{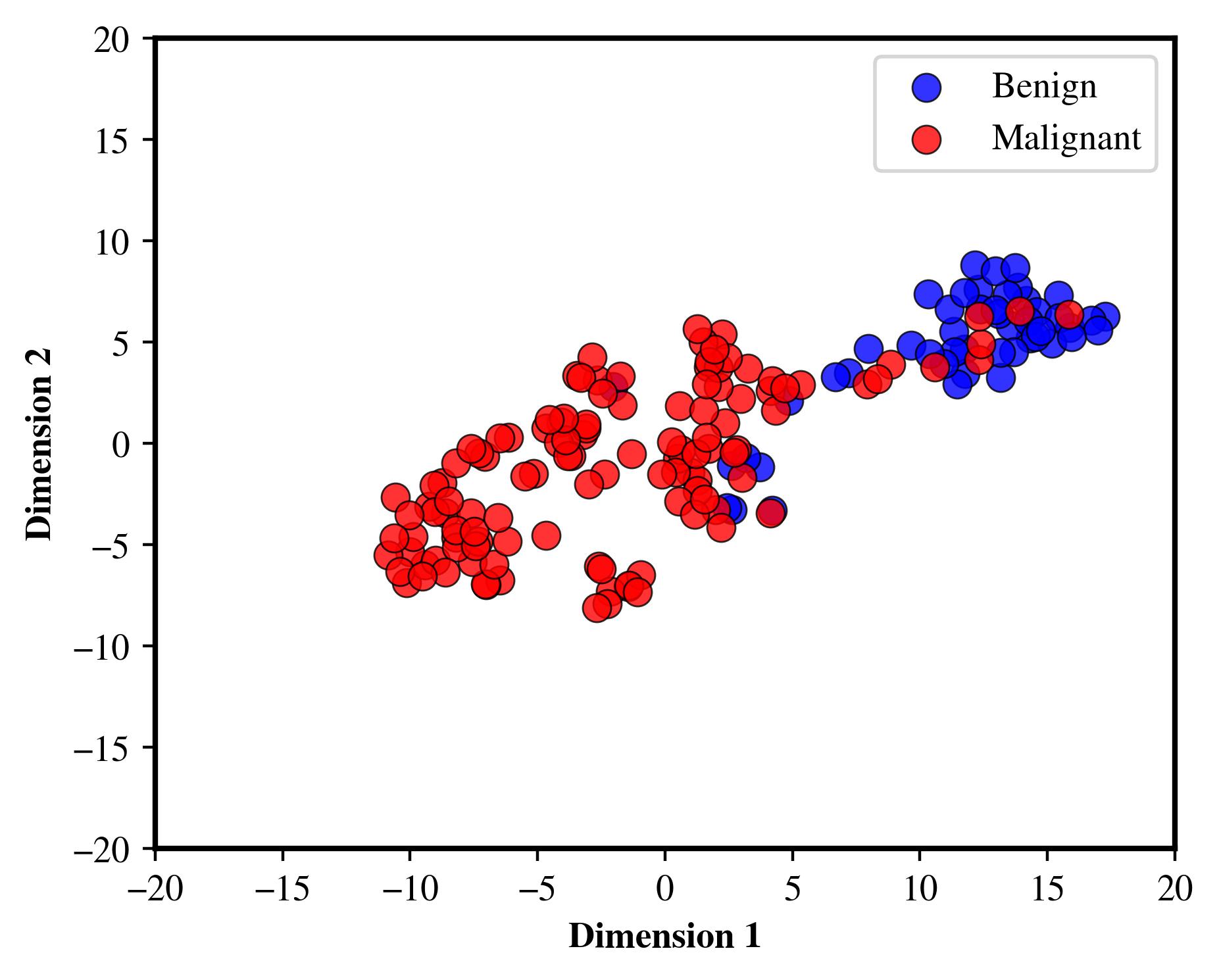}
    \end{minipage}
     \\ 
   \small (a) & \small (b) \\
    \end{tabular}
      \vspace{-3mm}
  \caption{t-SNE feature visualizations for the GNN-ViTCap for the BreakHis test dataset. (a) ResNet-34+DEC+GNN-MIL, (b) ViT+DEC+GNN-MIL. }
  \label{fig-t_sne}
\end{figure}

Table \ref{cls_com} demonstrates the performance of the proposed GNN-ViTCap and other SOTA methods for benign and malignant tumor classification on the BreakHis test dataset. In this experiment, the visual features of microscopic images are extracted using either a pretrained ResNet-34 or ViT-B/16 model. The SOTA MIL-based methods are conducted with the same configuration of the BreakHis test dataset to ensure a fair comparison. Overall, the proposed GNN-ViTCap ($ViT+ DEC+GNN-MIL$) architecture achieved the best classification performance across all metrics: $F_1-$Score of $0.934$, Precision of $0.926$, Recall of $0.942$, and AUC of $0.963$. The findings demonstrate an improvement of $2.3\%$ in the $F_1-$Score and $5.7\%$ in AUC over other SOTA MIL-based methods for classification, as shown in Table \ref{cls_com}. The SOTA MIL-based methods, including ABMIL \cite{ilse2018attention}, DSMIL \cite{li2021dual}, TransMIL \cite{shao2021transmil}, and DTFD-MIL \cite{zhang2022dtfd}, were designed for scanner-based whole slide images where the absolute positions of the patches are unknown. Therefore, the existing MIL-based methods show lower performance compared to the proposed GNN-ViTCap ($ViT+ DEC+GNN-MIL$) when using the microscopic images dataset. The reason behind that the proposed GNN-ViTCap removes redundant images or regions using a deep embedded clustering method and selects the most representative image patches based on attention score. Moreover, the GNN-based MIL identifies neighboring representative images based on their relative positions and aggregates the features for classification. Fig. \ref{fig-t_sne} depicts the feature distribution from the GNN-ViTCap method on the BreakHis test dataset using t-SNE visualization. 

\begin{table*}[!ht]
\centering
\caption{Performance metrics of the proposed GNN-ViTCap against SOTA methods for caption generation on PatchGastric test dataset.}
\vspace{-2mm}
\label{capresult} 
\renewcommand{\arraystretch}{1.25}
\begin{tabular}{ l l l l l l p{1.1cm} p{1cm} p{.8cm} p{.8cm} }
\hline
Methods & Visual Encoder & Language Model & BLEU-1 & BLEU-2 & BLEU-3& BLEU@4& METEOR& ROUGE& CIDEr\\ 
\hline 
\multirow{4}{*}{PatchCap \cite{tsuneki2022inference} }& EfficientNetB3\_AvgP &  \multirow{4}{*}{LSTM}   & -- & -- & -- & $0.283$ & $0.305$ &$ 0.497 $& $2.31$\\
& EfficientNetB3\_MaxP &   & -- & -- & -- &  $0.324$ & $0.277$ & $0.465$ &  $1.62$\\
& DenseNet121\_AvgP &  & -- & -- & -- &  $0.272$ & $0.282$&  $0.469$ & $1.52$ \\
& DenseNet121\_MaxP & & -- & -- & -- &  $0.323 $& $0.268$ & $0.454$ & $1.48 $\\
\hline
PathM3 \cite{zhou2024pathm3}& ViT-g/14 & Flan-T5& -- & -- & -- &$ 0.520 $& $0.394$ & $-$ & $-$ \\
\hline
SGMT \cite{qin2023whole} & CNN & Transformer& -- & -- & -- & $0.551$ & $0.432$ & $0.697$ &$ 4.83 $\\
\hline
\multirow{4}{*}{GNN-ViTCap}  & \multirow{4}{*}{ViT-B/16}  & BioGPT & $0.802$ & $0.748$ &$ 0.713$& $0.686$ & $0.485$ & $0.766$ & $5.72$\\
&  &ClinicalT5-Base & $0.851$ &$0.804$ & $0.774 $& $0.753 $& $0.526 $& $0.826$ & $6.72$\\
&  & LLamaV2-Chat & $0.877$ &$ 0.838 $&$ 0.813$ & $0.796$ & $0.557$ & $0.856$ & $7.25$\\
&  & BiomedGPT& \textbf{0.886} & \textbf{0.851} & \textbf{0.828} & \textbf{0.811} & \textbf{0.567} & \textbf{0.865} & \textbf{7.42}\\
\hline
\end{tabular}
\end{table*}

\begin{table*}[ht!]
\centering
  \vspace{-1mm}
\caption{Illustration of qualitative results using the GNN-ViTCap architecture on the PatchGastric test dataset} 
\vspace{-2mm}
\label{q_Result}
\renewcommand{\arraystretch}{1.2}
\begin{tabular}{ >{\centering\arraybackslash}c |  p{4cm} |   p{4cm} | l} 
\hline
Microscopic WSI & Ground Truth & GNN-ViTCap & BLEU@4 \\
\hline
\parbox[t]{2.2cm}{%
  \vspace{-1mm} 
  \hspace{0mm} 
  \includegraphics[width=2.2cm,height=1.5cm]{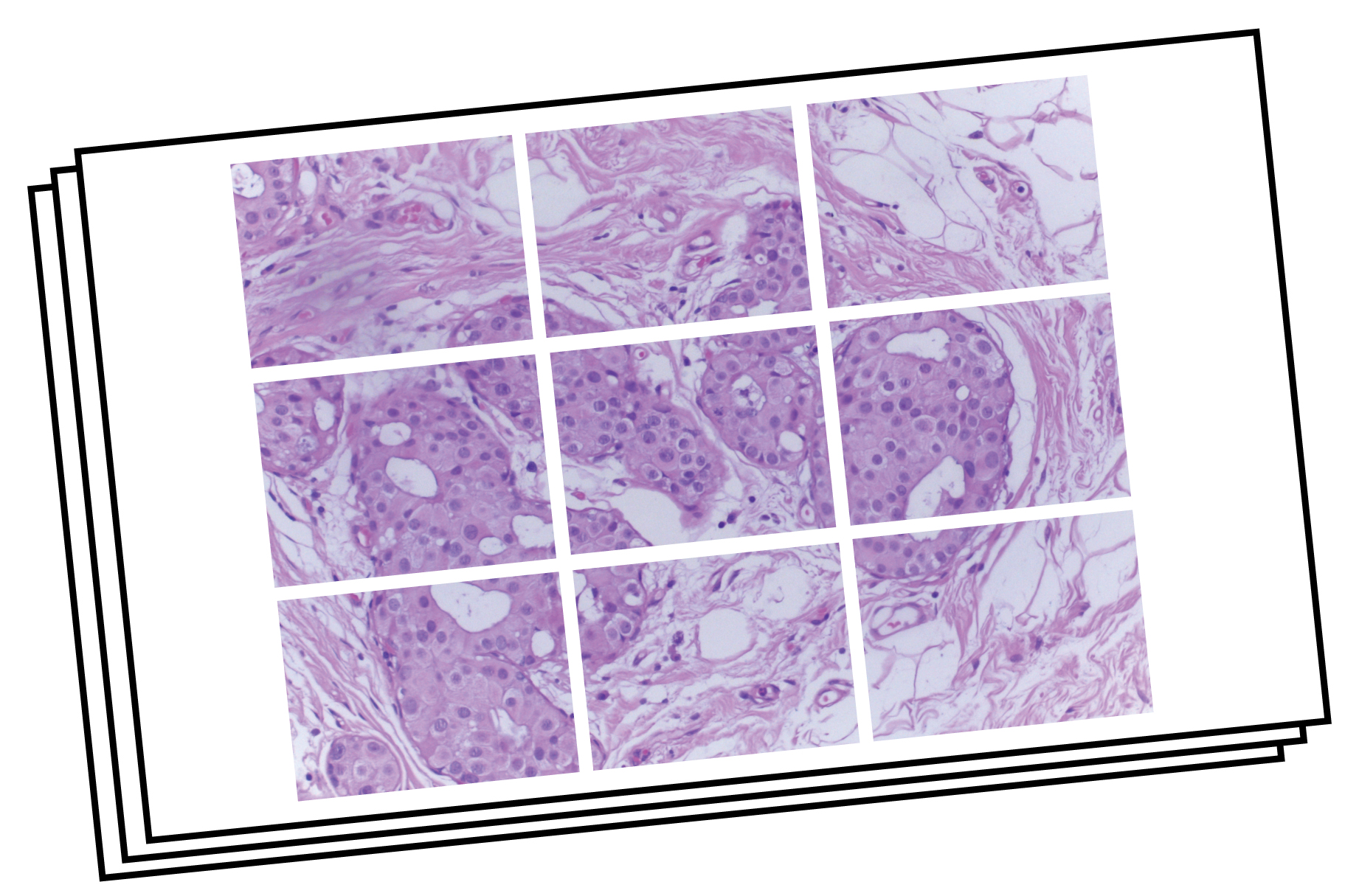}
}& \footnotesize tumor tissue in which medium to small irregular ducts infiltrate and proliferate in the submucosa can be irregular in the epithelium well differentiated tubular adenocarcinoma 
& \footnotesize tumor tissue in which medium to small irregular ducts infiltrate and proliferate in the submucosa can be seen in the epithelium well differentiated tubular adenocarcinoma \vspace{-3px} & \parbox[t]{1.2cm}{ \vspace{5mm}   \hspace{0mm} 0.889}  \\
\midrule
\parbox[t]{2.2cm}{%
  \vspace{-3mm} 
  \hspace{0mm} 
  \includegraphics[width=2.2cm,height=1.5cm]{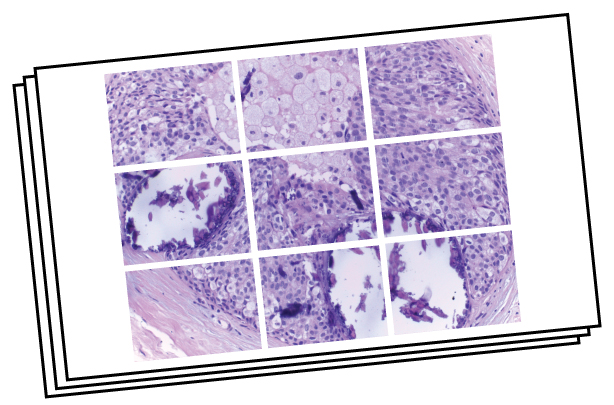}
}& \footnotesize \vspace{-10px} in the superficial epithelium tumor tissue that invades by forming medium sized to small irregular ducts is observed moderately differentiated adenocarcinoma
& \footnotesize  \vspace{-10px}  on the superficial epithelium tumor tissue that infiltrates by forming medium sized to small irregular ducts is observed moderately differentiated tubular &  \parbox[t]{1.2cm}{ \vspace{5mm}   \hspace{0mm} 0.768} \\
\hline
\end{tabular}
\end{table*}
\subsection{Image Captioning Results}
Table \ref{capresult} demonstrates the comparative analysis of the proposed GNN-ViTCap and other SOTA methods for image captioning on the PatchGastric test dataset. In the experiments, four large language models (ClinicalT5-Base, BioGPT, LLamaV2-Chat, and BiomedGPT) are fine-tuned using visual features along with corresponding captions. The proposed GNN-ViTCap ($ViT+GNN-MIL+BiomedGPT$) architecture achieved the highest scores across all metrics: a BLEU@4 score of $0.811$, METEOR of $0.567$, ROUGE of $0.865$, and CIDEr of $7.42$. The results indicate that our proposed GNN-ViTCap ($ViT+GNN-MIL+BiomedGPT$) method significantly outperforms the other SOTA methods in all metrics. The findings also demonstrate an improvement of $26\%$ on BLEU@4 and $13.5\%$ on METEOR over existing caption generation methods, as evident in Table \ref{capresult}. On the same dataset, the second-best BLEU@4 score of $0.796$ and METEOR of $0.557$ are obtained from the GNN-ViTCap ($ViT+GNN-MIL+LLamaV2-Chat$) architecture. The SOTA approach, PatchCap \cite{tsuneki2022inference} obtained the highest BLUE@4 score of $0.324$ using EfficientNetB3 and LSTM models. The other SOTA approaches, PathM3 \cite{zhou2024pathm3} and SGMT \cite{qin2023whole} achieved BLEU@4 of 0.520 and 0.551, respectively, using transformer-based language models. The superior performance of the GNN-ViTCap architecture is attributed to its effective feature extraction using a visual encoder and the selection of the most significant visual features through deep embedded clustering. Moreover, our proposed architecture explored the GNN-MIL, which aggregates features with patch positional encoding, and the integration with in-domain large language models provides a robust and effective solution for caption generation. Table \ref{q_Result} presents qualitative examples of the PatchGastric dataset by comparing generated captions from the proposed GNN-ViTCap method with reference captions.

\subsection{Discussions}
In this work, a novel GNN-ViTCap architecture is proposed for cancer tumor classification and caption generation from microscopic whole slide images. GNN-ViTCap architecture comprises a visual feature extractor, attention-based deep embedded cluster, GNN-based MIL, and large language models. Experimental results on both BreakHis and PatchGastric datasets demonstrate the effectiveness of the proposed GNN-ViTCap.

\textbf{\textit{\boldmath${Q_1}$: Does the proposed GNN-MIL perform better than SOTA MIL methods for microscopic WSI classification?}} The SOTA MIL-based methods, including ABMIL \cite{ilse2018attention}, DSMIL \cite{li2021dual}, TransMIL \cite{shao2021transmil}, and DTFD-MIL \cite{zhang2022dtfd}, were designed exclusively for scanner-based WSIs, whereas spatial information is absent in microscopic WSIs. However, the spatial information of patches in whole slide images is crucial for cancer diagnosis. Therefore, the proposed GNN-ViTCap ($ViT/ResNet-34 + DEC+ GNN-MIL$) learns spatial information from graph data. In GNN-ViTCap, each graph node represents a WSI patch, and edges are determined by the embedded features of these patches, capturing the spatial relationships between different regions. As a result, GNN-ViTCap efficiently captures both spatial information and patch correlations, leading to superior feature representations. Moreover, the GNN-ViTCap architecture removes redundant images or regions using a deep embedded clustering method and selects the most representative image patches based on attention scores. Therefore, the proposed GNN-ViTCap method outperforms the SOTA MIL-based methods for classification.

\textbf{\textit{\boldmath${Q_2}$: Does the spatial positional information of patches impact the performance of the model for caption generation?}}
The proposed GNN-ViTCap ($ViT+GNN-MIL+LLMs$) method achieved better results than the SGMT \cite{qin2023whole}, method for caption generation using the PatchGastric dataset. Both GNN-ViTCap and SGMT utilize transformer-based language models, but SGMT achieved a lower BLEU@4 score due to lack of consideration for patch positional encoding. The graph-based aggregation (GNN-MIL) within GNN-ViTCap leverages the spatial relationships between image patches, enabling the model to generate more accurate and relevant captions.

\textbf{\textit{\boldmath${Q_3}$: Do LLMs perform better than LSTM or traditional transformer models for image captioning of WSI?}}
Furthermore, the proposed GNN-ViTCap ($ViT+GNN-MIL+LLMs$) outperforms LSTM \cite{tsuneki2022inference} and traditional transformer models \cite{qin2023whole} in image captioning of whole slide images. Large language models have superior contextual understanding and efficient integration of multimodal data, which enable them to generate more accurate and coherent captions. The integration of advanced LLMs within the GNN-ViTCap framework facilitates more coherent and contextually accurate captioning.

\textbf{\textit{\boldmath${Q_4}$: Do in-domain LLMs perform better than generalized LLMs for generating captions in histopathological image analysis?}} 
Moreover, PathM3 \cite{zhou2024pathm3} introduced a multi-modal, multi-task, and multiple instance learning model for caption generation using ViT feature extraction and distilled versions of LLM models (Flan-T5). However, PathM3 obtained a lower BLEU@4 score compared to proposed GNN-ViTCap ($ViT+GNN-MIL+BimedGPT$) method, which utilizes in-domain LLMs. Fine-tuning language models on domain-specific data enables them to produce more relevant and precise descriptions, tailored to the nuances of medical imaging.


\section{Conclusions and Future Work}
In this paper, a novel GNN-ViTCap architecture is proposed for classification and caption generation from microscopic images. The GNN-ViTCap method is based on a visual feature extractor, attention-based deep embedded clustering, GNN-MIL aggregation, and LLMs. The deep embedded clustering method dynamically clusters images to reduce redundancy, while self-attention extracts the most representative images. Graph-based aggregation (GNN-MIL) leverages the spatial relationships between image patches and captures the contextual information. Therefore, LLMs are used for caption generation due to their exceptional context-association capabilities. Our proposed method is validated using the BreakHis and PatchGastric datasets. Experimental results demonstrate the method's effectiveness in microscopic image classification and captioning, aiding medical interpretation.

One of the major drawbacks of our proposed method is the sensitivity of the clustering method to the choice of $K$ value, which can lead to a high chance of information loss. In addition, fine-tuning the full LLMs is computationally expensive. In the future, adaptive clustering techniques can be explored to minimize information loss, along with parameter-efficient fine-tuning approaches to reduce the computational overhead of LLMs.


\balance
\bibliographystyle{IEEEtran} 
\bibliography{references}    


\end{document}